\documentclass{article}

\usepackage[final,nonatbib]{nips_2016}

\usepackage[utf8]{inputenc} % allow utf-8 input
\usepackage[T1]{fontenc}    % use 8-bit T1 fonts
\usepackage{url}            % simple URL typesetting
\usepackage{booktabs}       % professional-quality tables
\usepackage{amsfonts}       % blackboard math symbols
\usepackage{nicefrac}       % compact symbols for 1/2, etc.
\usepackage{microtype}      % microtypography

\usepackage{amsmath}
\usepackage{graphicx}
\usepackage{verbatim}
\usepackage{algorithm}
\usepackage{algorithmic}
\usepackage{color}
\usepackage{wrapfig}
\usepackage[bookmarks=false]{hyperref}

\long\def\ignorethis#1{}

\renewcommand{\eqref}[1]{Equation~(\ref{#1})}

\newcommand{\atari}{acvpa-oglls-15}
\newcommand{\nyu}{vpbmse-mcl-16}

\DeclareMathSymbol{@}{\mathord}{letters}{"3B}

\title{Unsupervised Learning for Physical Interaction through Video Prediction}

\author{
    Chelsea Finn\thanks{Work was done while the author was at Google Brain.}\\
  UC Berkeley\\
  \texttt{cbfinn@eecs.berkeley.edu}\\
  \And
  Ian Goodfellow\\
  OpenAI\\
  \texttt{ian@openai.com}\\
  \And
  Sergey Levine\\
  Google Brain\\
  UC Berkeley\\
  \texttt{slevine@google.com}\\
}
\begin{document}

\maketitle

\begin{abstract}

A core challenge for an agent learning to interact with the world is to predict how its actions affect objects in its environment.
Many existing methods for learning the dynamics of physical interactions require labeled object
information. However, to scale real-world interaction learning
to a variety of scenes and objects, acquiring labeled data becomes increasingly impractical.
To learn about physical object motion without labels, we develop an action-conditioned video
prediction model that explicitly models pixel motion, by predicting a distribution over pixel motion from previous frames.
Because our model explicitly predicts motion, it is partially invariant to object appearance, enabling it to generalize to previously unseen objects.
To explore video prediction for real-world interactive agents, we also introduce a dataset of $59@000$ robot interactions involving
pushing motions, including a test set with novel objects. In this dataset, accurate prediction of videos conditioned on the robot's future actions
amounts to learning a ``visual imagination'' of different futures based on different courses of action.
Our experiments show that our proposed method produces more accurate
video predictions both quantitatively and qualitatively, when compared to prior methods.

\end{abstract}

\section{Introduction}
\label{intro}

Object detection, tracking, and motion prediction are fundamental problems in computer vision, and
predicting the effect of physical interactions is a critical challenge for learning agents acting in the world, such as robots, autonomous cars,
and drones. % other examples? simulated agents?
Most existing techniques for learning to predict physics rely on large manually labeled datasets (e.g. \cite{mbrf-niu-15}).
However, if interactive agents can use unlabeled raw video data to learn about physical interaction, they can autonomously collect virtually unlimited experience through their own exploration.
%Even without labels, raw video provides a rich source of information, showing how objects physically move across the background, become occluded, and respond to actions taken by the agent.
Learning a representation which can predict future video without labels has applications in action recognition and prediction and, when conditioned on the action of the agent,
amounts to learning a predictive model that can then be used for planning and decision making.

%visual model-based reinforcement learning (RL).

However, learning to predict physical phenomena poses many challenges, since real-world physical interactions tend to be complex and stochastic,
and learning from raw video requires handling the high dimensionality of image pixels and the partial observability of object motion from videos.
Prior video prediction methods have typically considered short-range prediction \cite{\nyu}, small image patches \cite{ulvr-sms-15},
or synthetic images~\cite{\atari}.
Such models follow a paradigm of reconstructing future frames from the internal state of the model. In our approach, we propose a method which does not require the model
to store the object and background appearance. Such appearance information is directly available in the previous frame. We develop a predictive model which merges
appearance information from previous frames with motion predicted by the model. As a result, the model is better able to predict future video sequences for multiple steps,
even involving objects not seen at training time.

% paragraph on the model
To merge appearance and predicted motion, we output the motion of pixels relative to the previous image. Applying this motion to the previous image forms the next frame.
We present and evaluate three motion prediction modules. The first, which we refer to as dynamic neural advection (DNA), outputs a distribution over locations in the previous frame for each pixel in the new frame.
The predicted pixel value is then computed as an expectation under this distribution.
A variant on this approach, which we call convolutional dynamic neural advection (CDNA), outputs the parameters of multiple
normalized convolution kernels to apply to the previous image to compute new pixel values.
The last approach, which we call spatial transformer predictors (STP),
outputs the parameters of multiple
affine transformations to apply to the previous image, akin to the spatial transformer network
previously proposed for supervised learning~\cite{jsz-stn-15}.
In the case of the latter two methods, each predicted transformation is meant to handle separate objects. To combine the predictions into a single image, the model also predicts a compositing mask over each of the
transformations. DNA and CDNA are simpler and easier to implement than STP, and while all models achieve comparable performance,
the object-centric CDNA and STP models also provide interpretable internal representations.
%%SL.5.19b: I edited the preceding sentence, perhaps a little ambitiously. It would be really good if we can do something like that.
% this might change with flow/motion results.

% comment on conv lstm? (doesn't fit particularly well here)
%Because physics is consistent across different parts of the image, we also employ
%convolutional LSTMs.
%%SL.5.16: this last part could use more motivation.

Our main contribution is a method for making long-range predictions in real-world videos
by predicting pixel motion. When conditioned on the actions
taken by an agent, the model can learn to imagine different futures from different actions.
To learn about physical interaction from videos, we need a large dataset with complex object interactions.
We collected a dataset of $59@000$ robot pushing motions, consisting of $1.5$ million frames
and the corresponding actions at each time step.
Our experiments using this new robotic pushing dataset, and using a human motion video dataset~\cite{ipos-h36m-14},
show that models that explicitly
transform pixels from previous frames better capture object motion and produce more accurate video predictions compared to prior state-of-the-art methods.
The dataset, video results, and code are all available online: \url{sites.google.com/site/robotprediction}.

\section{Related Work}
\label{related}

%\todo{this section might be better organized without paragraph structure}

%%SL.5.20: I commented this out for length, I think nothing is lost.
%Our contributions are related to prior work on video prediction, learning physics, unsupervised learning for interaction, and video data collection.
% main camps: video prediction models (but not used for control), physics learning models (require labels, or simplistic), unsupervised learning for control (trained on recon), datasets (missing niche)

\vspace{-0.07in}
\paragraph{Video prediction:} Prior work on video prediction has tackled synthetic videos and short-term prediction in real videos. Yuan et al.~\cite{Yuen2010} used a
nearest neighbor approach to construct predictions from similar videos in a dataset.
Ranzato et al. proposed a baseline for video prediction inspired by language models \cite{vlmb-rsbmcc-14}.
LSTM models have been adapted for video prediction on patches~\cite{ulvr-sms-15}, action-conditioned Atari frame predictions~\cite{acvpa-oglls-15}, and precipitation
nowcasting \cite{xcwyw-clstm-15}.
Mathieu et al. proposed new loss functions for sharper frame predictions~\cite{vpbmse-mcl-16}.
Prior methods generally reconstruct frames from the internal state of the model, and some predict the internal state directly, without producing images~\cite{vpt-afwuv-15}.
Our method instead transforms pixels from previous frames, explicitly modeling motion and, in the case of the CDNA and STP models, decomposing it over image segments.
We found in our experiments that all three of our models produce substantially better predictions by advecting pixels from the previous frame and compositing them onto
the new image, rather than constructing images from scratch.
This approach differs from recent work on optic flow prediction~\cite{wdgh-auf-16}, which predicts where pixels will move to using direct optical flow supervision.
Boots et al. predict future images of a robot arm using nonparametric kernel-based methods \cite{bbf-lpmdc-14}.
In contrast to this work, our approach uses flexible parametric models,
and effectively predicts interactions with objects, including objects not seen during training. To our knowledge, no previous video prediction method has been applied to predict real images with novel object interactions beyond two time steps into the future.

There have been a number of promising methods for frame prediction developed concurrently to this work~\cite{lkc-dpcn-16}.
Vondrick et al.~\cite{vpt-gcsd-16} combine an adversarial objective with a multiscale, feedforward architecture, and use a foreground/background mask similar
to the masking scheme proposed here.
De Brabandere et al.~\cite{djtv-dfn-16} propose a method similar to our DNA model, but use a softmax for sharper flow distributions.
The probabilistic model proposed by Xue et al.~\cite{xwbf-vd-16} predicts transformations applied to latent feature maps, rather than the image itself, but only demonstrates
single frame prediction.

\vspace{-0.07in}
\paragraph{Learning physics:} Several works have explicitly addressed prediction of physical interactions, including
predicting ball motion \cite{bsf-ecd-09}, block falling \cite{bht-sepse-13},
the effects of forces
\cite{mrgf-whi-16,mbrf-niu-15}, future human interactions \cite{hk-ar-14}, and future car trajectories \cite{wgh-pttf-14}. These methods require ground truth object pose information, segmentation masks, camera viewpoint, or image patch trackers. In the domain of reinforcement learning, model-based methods have been proposed that learn prediction on images \cite{lrv-arl-12,e2c-wsbr-15}, but they have either used synthetic images or instance-level models, and have not demonstrated generalization to novel objects nor accurate prediction on real-world videos. As shown by our comparison to LSTM-based prediction designed for Atari frames \cite{acvpa-oglls-15}, models that work well on synthetic domains do not necessarily succeed on real images.

\vspace{-0.07in}
\paragraph{Video datasets:}
Existing video datasets capture YouTube clips~\cite{ktsls-lsvc-14}, human motion~\cite{ipos-h36m-14},
synthetic video game frames~\cite{\atari}, and driving~\cite{glsu-kitti-13}.
However, to investigate learning visual physics prediction,
we need data that exhibits rich object motion, collisions, and interaction information. We propose a large new dataset consisting of real-world
videos of robot-object interactions, including complex physical phenomena, realistic occlusions, and a clear use-case for interactive robot learning.

%%SL.5.5: there are probably other non-deep-learning papers
%Different areas of related work (not covered above yet):
%\begin{itemize}
%\item capsules \cite{ta-hkw-11}, object discovery/tracking \cite{kclps-uodt-15}
%\item Abhinav unsupervised learning from video ICCV paper \cite{ulvrv-wg-15}
%\end{itemize}

\section{Motion-Focused Predictive Models}
\label{sec:models}

\begin{figure}
	\setlength{\unitlength}{0.5\columnwidth}
	\begin{picture}(1.99,0.7) \linethickness{0.5pt}
	\put(0.0,0.0){\includegraphics[width=1.0\columnwidth]{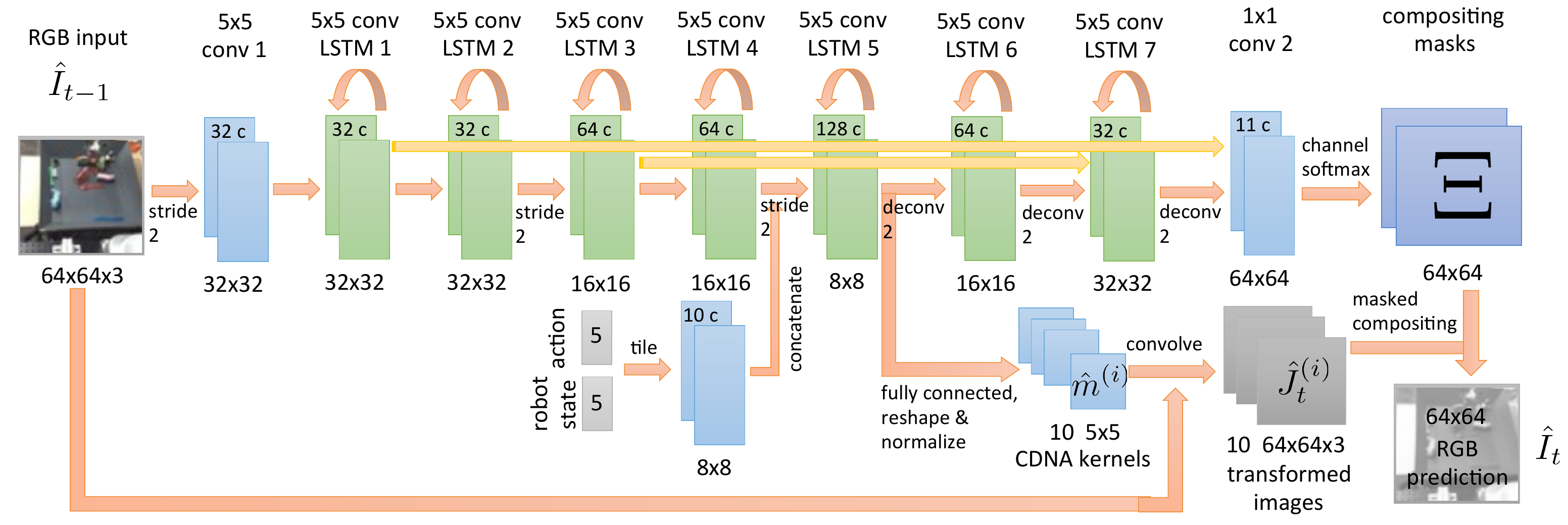}}
	\end{picture}
	\caption{Architecture of the CDNA model, one of the three proposed pixel advection models. We use convolutional LSTMs to process the image, outputting $10$ normalized transformation kernels from the smallest middle layer of the network and an
	11-channel compositing mask from the last layer (including $1$ channel for static background). The kernels are applied to transform the previous image into $10$ different transformed images,
	which are then composited according to the masks. The masks sum to $1$ at each pixel due to a channel-wise softmax. Yellow arrows denote skip connections.
		\label{fig:diagram}
	}
	\vspace{-0.1in}
\end{figure}

In order to learn about object motion while remaining invariant to appearance,
we introduce a class of video prediction models that directly use appearance
information from previous frames to construct pixel predictions. Our model computes the next frame by first predicting the motions of
image segments, then merges these predictions via masking. In this section, we discuss our novel pixel transformation models,
and propose how to effectively merge predicted motion of multiple segments into a single next image prediction.
The architecture of the CDNA model is shown in Figure~\ref{fig:diagram}. Diagrams of the DNA and STP models are in Appendix~\ref{app:model}.

%Lastly, we discuss an efficient architecture for making such motion predictions.
%\todo{refer to figure}

\subsection{Pixel Transformations for Future Video Prediction}
The core of our models is a motion prediction module that predicts objects' motion without attempting to reconstruct
their appearance. This module is therefore partially invariant to appearance and can generalize effectively to previously unseen objects.
We propose three motion prediction modules:

%Because we are concerned with learning physics, the model should not be required to reconstruct object appearance purely
%from its own internal representation, especially in the presence of irrelevant distractors such as lighting and background variation.
%Instead of requiring the network to reconstruct object appearances, we develop a model which reuses appearance information in previous frames
%and focuses its representational power entirely on predicting physics. We propose three motion prediction modules:

\vspace{-0.07in}
\paragraph{Dynamic Neural Advection (DNA):} In this approach, we predict a distribution over locations in the previous frame for each pixel in the
new frame. The predicted pixel value is computed as an expectation under this distribution.
We constrain the pixel movement to a local region, under the regularizing assumption that
pixels will not move large distances. This keeps the dimensionality of the prediction low. This approach is the most flexible of the proposed approaches.

Formally, we apply the predicted motion transformation $\hat{m}$ to the previous image prediction $\hat{I}_{t-1}$ for every pixel $(x,y)$ to form the next image prediction $\hat{I}_t$ as follows:
$$
\hat{I}_t(x,y) = \sum_{k \in (-\kappa,\kappa)} \sum_{l \in (-\kappa, \kappa)}  \hat{m}_{xy}(k,l) \hat{I}_{t-1}(x-k,y-l)
$$

where $\kappa$ is the spatial extent of the predicted distribution. This can be implemented as a convolution with untied weights.
The architecture of this model matches the CDNA model in Figure~\ref{fig:diagram},
except that the higher-dimensional transformation parameters $\hat{m}$ are outputted by the last (conv 2) layer instead of the LSTM 5 layer used for the CDNA model.
%At the first prediction step, the previous image is a true image.

% not sure if I like this first sentence
\vspace{-0.07in}
\paragraph{Convolutional Dynamic Neural Advection (CDNA):} Under the assumption that the same mechanisms can be used to predict the motions of
different objects in different regions of the image, we consider a more object-centric approach to predicting motion.
Instead of predicting a different
distribution for each pixel, this model predicts multiple discrete distributions that are each applied to the entire image via a convolution (with tied weights),
which computes the expected value of the motion distribution for every pixel.
The idea is that pixels on the same rigid object will move together, and therefore
can share the same transformation. More formally, one predicted object transformation $\hat{m}$ applied to the previous image $I_{t-1}$ produces image $\hat{J}_t$ for
each pixel $(x,y)$ as follows:
$$
\hat{J}_t(x,y) = \sum_{k \in (-\kappa,\kappa)} \sum_{l \in (-\kappa, \kappa)} \hat{m}(k,l) \hat{I}_{t-1}(x-k,y-l)
$$
where $\kappa$ is the spatial size of the normalized predicted convolution kernel $\hat{m}$. Multiple transformations $\{\hat{m}^{(i)}\}$ are applied to the
previous image $\hat{I}_{t-1}$ to form multiple images $\{\hat{J}_t^{(i)}\}$.
These output images are combined into a single prediction $\hat{I}_t$ as described in the next section and show in Figure~\ref{fig:diagram}.

\vspace{-0.07in}
\paragraph{Spatial Transformer Predictors (STP):}
In this approach, the model produces multiple sets of parameters for 2D affine image transformations, and applies the transformations using
a bilinear sampling kernel \cite{jsz-stn-15}. More formally, a set of affine parameters $\hat{M}$ produces a warping grid between previous image pixels $(x_{t-1},y_{t-1})$
and generated image pixels $(x_t,y_t)$.
$$
\begin{pmatrix}
    x_{t-1}\\
    y_{t-1}
\end{pmatrix}
=
\hat{M}
\begin{pmatrix}
    x_{t}\\
    y_{t}\\
    1
\end{pmatrix}
$$
This grid can be applied with a bilinear kernel to form an image $\hat{J}_t$:
\vspace{-0.1cm}
$$
\hat{J}_t(x_t,y_t) = \sum_k^W \sum_l^H \hat{I}_{t-1}(k,l) \max(0,1-|x_{t-1}-k|) \max(0, 1-|y_{t-1}-l|)
\vspace{-0.1cm}
$$
where $W$ and $H$ are the image width and height.
While this type of operator has been applied previously only to supervised learning tasks, it is well-suited for video prediction.
Multiple transformations $\{\hat{M}^{(i)}\}$ are applied to the previous image $\hat{I}_{t-1}$ to form multiple images $\{\hat{J}_t^{(i)}\}$,
which are then composited based on the masks. The architecture matches the diagram in Figure~\ref{fig:diagram},
but instead of outputting CDNA kernels at the LSTM 5 layer, the model outputs the STP parameters $\{\hat{M}^{(i)}\}$.

All of these models can
focus on learning physics rather than object appearance. Our experiments show that these models
are better able to generalize to unseen objects compared to models that reconstruct the pixels directly or predict the difference from the previous frame.

\subsection{Composing Object Motion Predictions}

CDNA and STP produce multiple object motion predictions, which need to be combined into a single image.
The composition of the predicted images $\hat{J}_{t}^{(i)}$ is modulated by a mask $\Xi$, which defines a weight on each prediction, for each pixel.
Thus,
%$\hat{I}_{i,j,t} = \sum_c \Xi_{i,j,c} \hat{J}^{(c)}_{i,j,t} $
$\hat{I}_t = \sum_c \hat{J}^{(c)}_t \circ \Xi_c$
, where $c$ denotes the channel of the mask and the element-wise multiplication is over pixels.
To obtain the mask, we apply a channel-wise softmax to the final convolutional layer in the model (conv 2 in Figure~\ref{fig:diagram}),
which ensures that the channels of the mask sum to 1 for each pixel position.

%To ensure that the channels of the masks $\Xi$ sum to one, $\Xi$ is generated via a channel-wise softmax applied to activations of the final convolutional layer.
%These masks indicate how much each transformed image influences each pixel.
%The masks $\Xi$ are formed via a channel-wise softmax on the final network activations, ensuring that each pixel value in $\hat{I}_t$ is determined by a convex combination of the corresponding pixels in the transformed images $\hat{J}_t$, as in the following equation:
%A softmax over the channels of the mask ensures that it sums to one.

%%SL.5.20: I think you need an equation to explain the softmax over channels, currently it's way too brief of a description.
In practice, our
experiments show that the CDNA and STP models learn to mask out objects that are moving in consistent directions.
The benefit of this approach is two-fold: first, predicted motion transformations are reused
for multiple pixels in the image, and second, the model naturally extracts a more object centric representation in an unsupervised fashion, a desirable
property for an agent learning to interact with objects. The DNA model lacks these two benefits, but instead is more flexible
as it can produce independent motions for every pixel in the image.

For each model, including DNA, we also include a ``background mask'' where we allow the models to copy pixels directly from the previous frame.
Besides improving performance, this also produces interpretable background masks that we visualize in Section~\ref{experiments}.
Additionally, to fill in previously occluded regions, which may not be well represented by nearby pixels, we allowed the models
to generate pixels from an image, and included it in the final masking step.

\subsection{Action-conditioned Convolutional LSTMs}

Most existing physics and video prediction models use feedforward architectures~\cite{\nyu,lgf-lpibt-16} or feedforward encodings of the image~\cite{\atari}.
To generate the motion predictions discussed above, we employ stacked convolutional LSTMs~\cite{xcwyw-clstm-15}.
Recurrence through convolutions is a natural fit for multi-step video prediction because it takes advantage of the spatial invariance of image representations, as
the laws of physics are mostly consistent across space.
As a result, models with convolutional recurrence require significantly fewer parameters and use those parameters more efficiently.

The model architecture is displayed in Figure~\ref{fig:diagram}
and detailed in Appendix~\ref{app:model}. In an interactive setting, the agent's actions and internal state (such as the pose of the robot gripper) influence the next
image. We integrate both into our model by spatially tiling the concatenated state and action vector across a feature map, and concatenating the result
to the channels of the lowest-dimensional activation map.
Note, though, that the agent's internal state (i.e. the robot gripper pose) is only input into the network at the beginning,
and must be predicted from the actions in future timesteps. We trained the networks using an $l_2$ reconstruction loss.
Alternative losses, such as those presented in~\cite{\nyu} could complement this method.

% rest of details in experiments

\section{Robotic Pushing Dataset}
\label{sec:dataset}

\begin{wrapfigure}{r}{0.4\textwidth}
\vspace{0.3cm}
\setlength{\unitlength}{0.5\columnwidth}
\begin{picture}(1.99,0.45) \linethickness{0.5pt}
\put(0.0,0.2){\includegraphics[width=0.4\columnwidth]{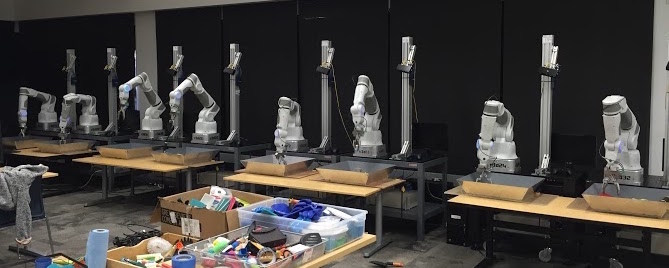}}
\put(0.0,-0.05){\includegraphics[width=0.12\columnwidth]{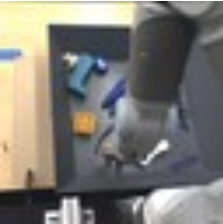}}
\put(0.28,-0.05){\includegraphics[width=0.12\columnwidth]{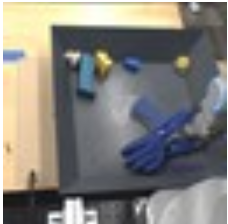}}
\put(0.56,-0.05){\includegraphics[width=0.12\columnwidth]{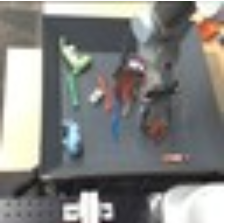}}
\end{picture}
\vspace{-0.07in}
\caption{Robot data collection setup (top) and example images captured from the robot's camera (bottom).
\label{fig:data}
}
\vspace{-0.10in}
\end{wrapfigure}

One key application of action-conditioned video prediction is to use the learned model for decision making in vision-based robotic control tasks.
Unsupervised learning from video can enable agents to learn about the world on their own, without human involvement, a critical requirement
for scaling up interactive learning. In order to investigate action-conditioned video prediction for robotic tasks, we need a dataset with real-world physical object interactions.
We collected a new dataset using $10$ robotic arms, shown in Figure \ref{fig:data}, pushing hundreds of
objects in bins, amounting to $57@000$ interaction sequences with $1.5$ million video frames.
Two test sets, each with $1@250$ recorded motions, were also collected. The first test set used two different subsets of the objects pushed during training.
The second test set involved two subsets of objects, none of which were used during training.
In addition to RGB images, we also record the corresponding gripper poses, which we refer to as the internal state,
and actions, which corresponded to the commanded gripper pose. The dataset is publically available\footnote{See \url{http://sites.google.com/site/robotprediction}}.
Further details on the data collection procedure are provided in Appendix~\ref{app:robot}.

\section{Experiments}
\label{experiments}

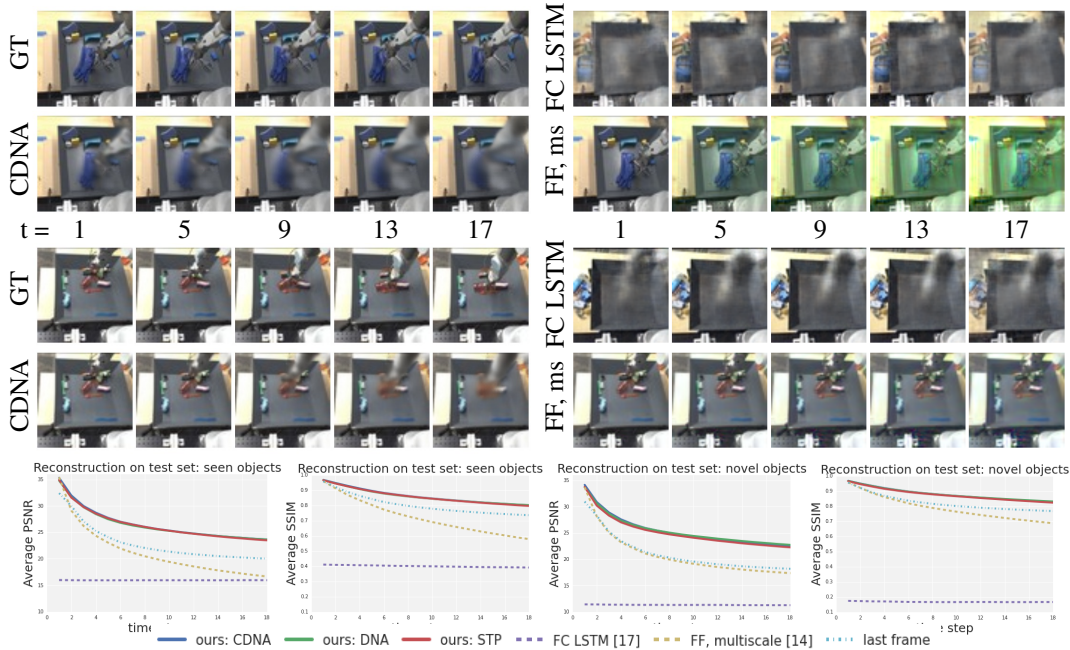
\begin{figure}
\setlength{\unitlength}{0.5\columnwidth}
\begin{picture}(1.99,1.15) \linethickness{0.5pt}
    \put(0.01,0.97){\includegraphics[width=0.48\columnwidth]{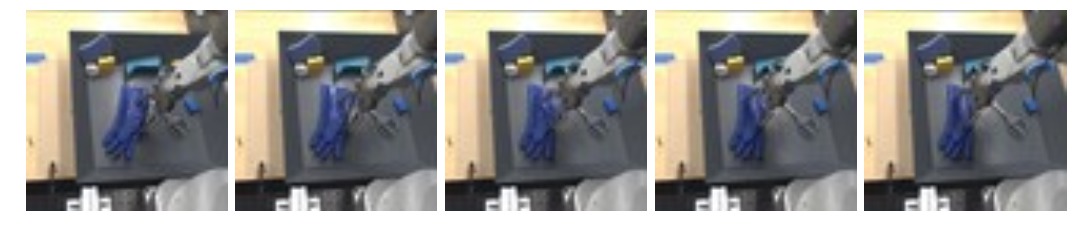}}
    \put(0.01,0.77){\includegraphics[width=0.48\columnwidth]{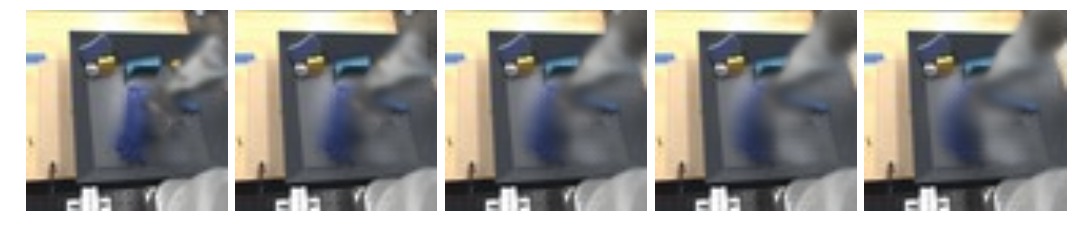}}
    \put(1.03,0.97){\includegraphics[width=0.48\columnwidth]{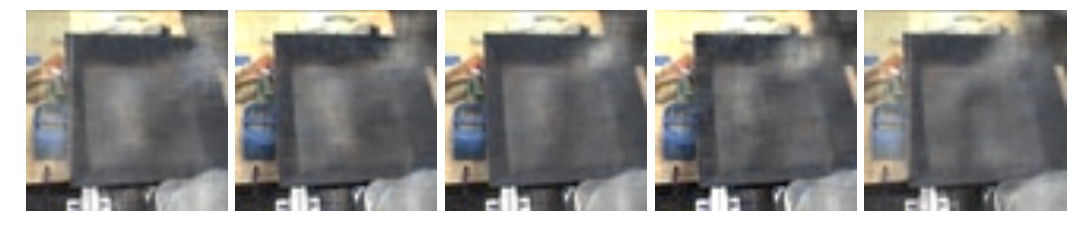}}
    \put(1.03,0.77){\includegraphics[width=0.48\columnwidth]{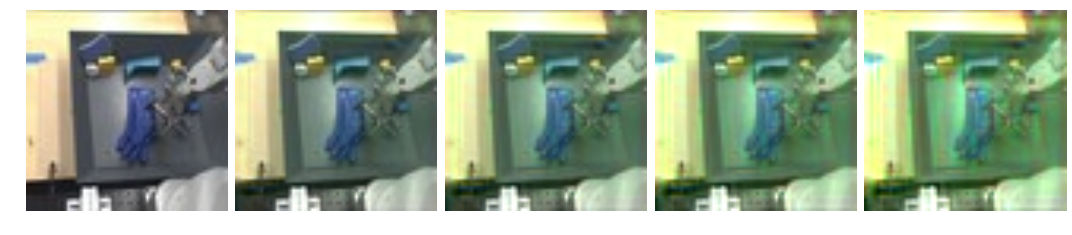}}
    \put(-0.02, 1.06){\rotatebox{90}{GT}}
    \put(-0.02, 0.82){\rotatebox{90}{CDNA}}
    \put(1.0, 0.98){\rotatebox{90}{FC LSTM}}  % 0.53
    \put(1.0, 0.82){\rotatebox{90}{FF, ms}}

    \put(0.01,0.52){\includegraphics[width=0.48\columnwidth]{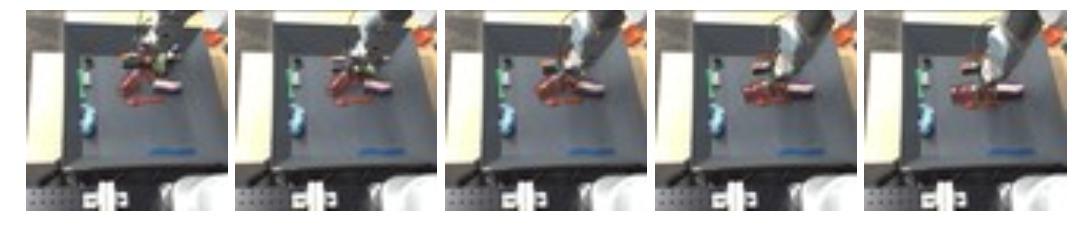}}
    \put(0.01,0.32){\includegraphics[width=0.48\columnwidth]{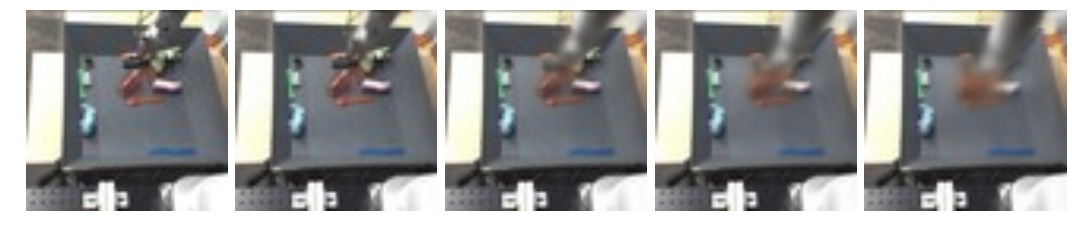}}
    \put(1.03,0.52){\includegraphics[width=0.48\columnwidth]{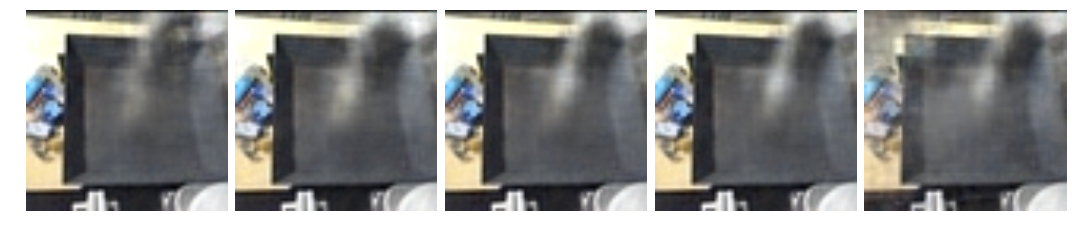}}
    \put(1.03,0.32){\includegraphics[width=0.48\columnwidth]{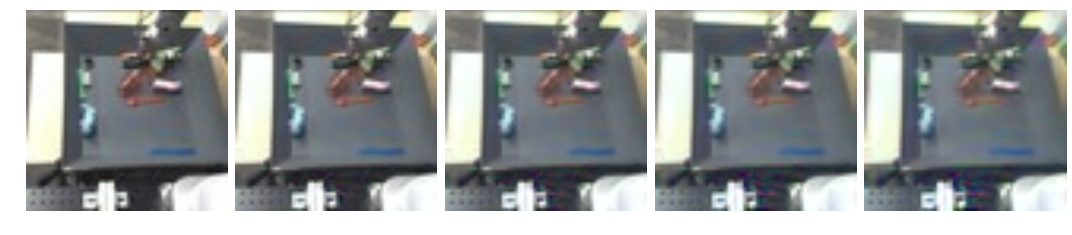}}

    \put(0.0,-0.02){\includegraphics[width=0.25\columnwidth]{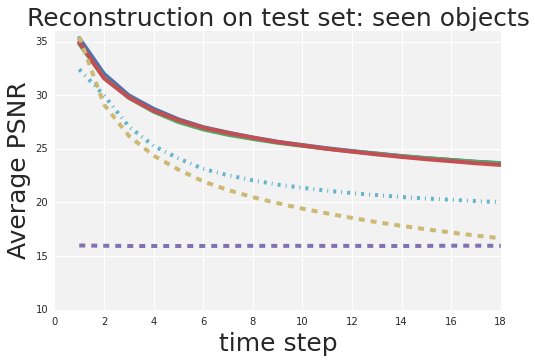}} % was 0.42
    \put(0.5,-0.02){\includegraphics[width=0.25\columnwidth]{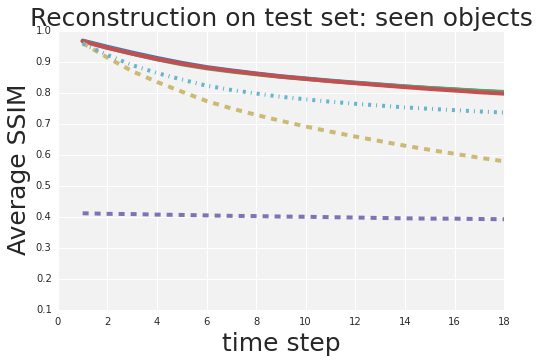}}
    %\put(1.33,0.42){\includegraphics[width=0.34\columnwidth]{flow_seen.png}}

    \put(1.0,-0.02){\includegraphics[width=0.25\columnwidth]{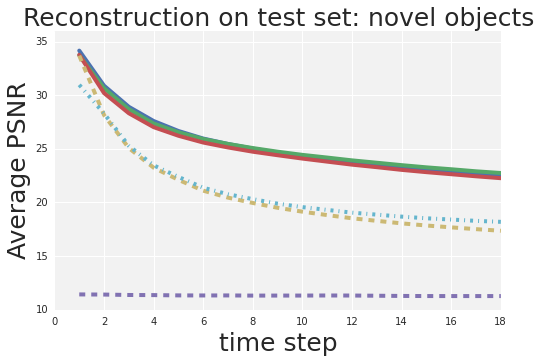}}
    \put(1.5,-0.02){\includegraphics[width=0.25\columnwidth]{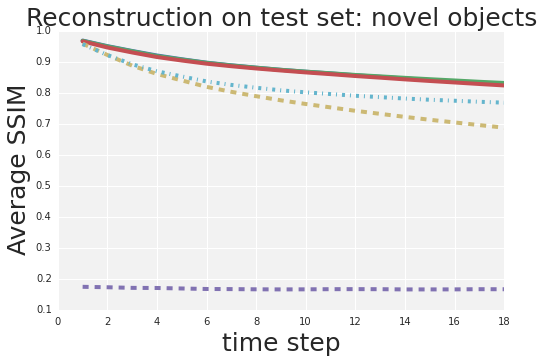}}
    %\put(1.33,-0.0){\includegraphics[width=0.34\columnwidth]{flow_novel.png}}

    \put(-0.02, 0.61){\rotatebox{90}{GT}}  % 0.53
    \put(-0.02, 0.37){\rotatebox{90}{CDNA}}
    \put(1.0, 0.53){\rotatebox{90}{FC LSTM}}
    \put(1.0, 0.37){\rotatebox{90}{FF, ms}}

    \put(0.0, 0.74){t =}
    \put(0.1, 0.74){1}
    \put(0.3, 0.74){5}
    \put(0.49, 0.74){9}
    \put(0.67, 0.74){13}
    \put(0.85, 0.74){17}

    \put(1.13, 0.74){1}
    \put(1.32, 0.74){5}
    \put(1.51, 0.74){9}
    \put(1.68, 0.74){13}
    \put(1.87, 0.74){17}
    \put(0.25,-0.05){\includegraphics[width=0.75\columnwidth]{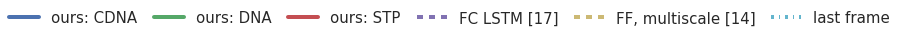}}

\end{picture}
\vspace{0.2cm}
\caption{Qualitative and quantitative reconstruction performance of our models, compared with \cite{\atari,\nyu}.
All models were trained for 8-step prediction, except \cite{\nyu}, trained for 1-step prediction.
\label{fig:videos}
}
\vspace{-0.1in}
\end{figure}

We evaluate our method using the dataset in Section~\ref{sec:dataset}, as well as on videos of human motion in
the Human3.6M dataset~\cite{ipos-h36m-14}.
In both settings, we evaluate our three models described in Section~\ref{sec:models}, as well as prior models \cite{\nyu,\atari}. For CDNA and STP, we
used $10$ transformers.
%For all three models, in addition to the transformations described, we also allow the network to copy background pixels directly from the
%previous frame by including the previous image in the masked compositing operation, with an additional channel in the mask for the background.
% should probably say something more high level here... but what?
%%SL.5.19: maybe something about the aims of the experiments would be good to state here
While we show stills from the predicted videos in the figures, the qualitative results are easiest to compare when the predicted videos can be viewed side-by-side.
For this reason, we encourage the reader to examine the video results on the supplemental website$^2$. Code for training the model
is also available on the website.

\vspace{-0.07in}
\paragraph{Training details:} We trained all models using the TensorFlow
library~\cite{aabbc-tf-15}, optimizing to convergence using ADAM~\cite{kb-adam-15} with the suggested hyperparameters.
We trained all recurrent models with and without scheduled sampling~\cite{bvjs-ss-15} and report the performance of the model with the best validation error.
We found that scheduled sampling improved performance of our models, but did not substantially affect the performance of ablation and baseline
models that did not model pixel motion.

\subsection{Action-conditioned prediction for robotic pushing}

Our primary evaluation is on video prediction using our robotic interaction dataset, conditioned on the future actions taken by the robot.
In this setting, we pass in two initial images, as well as the initial robot arm state and actions, and then sequentially roll out the model,
passing in the future actions and the model's image and state prediction from the previous time step.
We trained for $8$ future time steps for all recurrent models, and test for up to 18 time steps. We held out 5\% of
the training set for validation. To quantitatively evaluate the predictions, we measure average PSNR and SSIM, as proposed in~\cite{\nyu}.
Unlike~\cite{\nyu}, we measure these metrics on the entire image.
We evaluate on two test sets described in Section~\ref{sec:dataset}, one with objects seen at training time, and one with previously unseen objects.

Figure~\ref{fig:videos} illustrates the performance of our models compared to prior methods.
We report the performance of the feedforward multiscale model of \cite{\nyu} using an $l_1$+GDL loss, which was the best performing model in our experiments
-- full results of the multi-scale models are in Appendix~\ref{app:multiscale}.
Our methods significantly outperform prior video prediction methods on all metrics. The FC LSTM model \cite{\atari}
reconstructs the background and lacks the representational power to reconstruct the objects in the bin.
The feedforward multiscale model performs well on 1-step prediction, but performance quickly drops over time, as it is only trained for 1-step prediction.
It is worth noting that our models
are significantly more parameter efficient: despite being recurrent, they contain $12.5$ million parameters, which is slightly less than the feedforward model with
$12.6$ million parameters and significantly less than the FC LSTM model which has $78$ million parameters.
We found that none of the models suffered from significant overfitting on this dataset. We also report the baseline performance
of simply copying the last observed ground truth frame.
%However, based on validation error, neither our models nor the baseline models suffered from overfitting on this dataset.
% atari - 78 million parameters, nyu - 12.6 million parameters (yet only ff), ours 12.6 million parameters (and recurrent)/vis

\begin{figure}
\setlength{\unitlength}{0.5\columnwidth}
\begin{picture}(1.99,0.3) \linethickness{0.5pt}
    \put(0.0,0.0){\includegraphics[width=0.25\columnwidth]{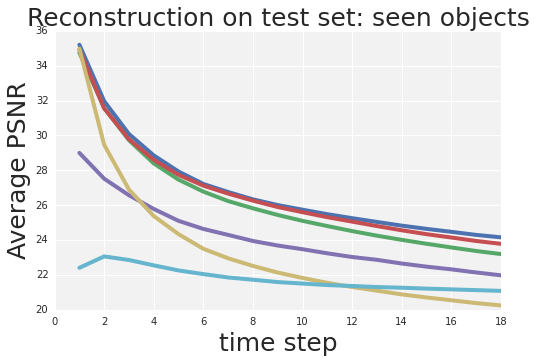}}
    \put(0.5,0.0){\includegraphics[width=0.25\columnwidth]{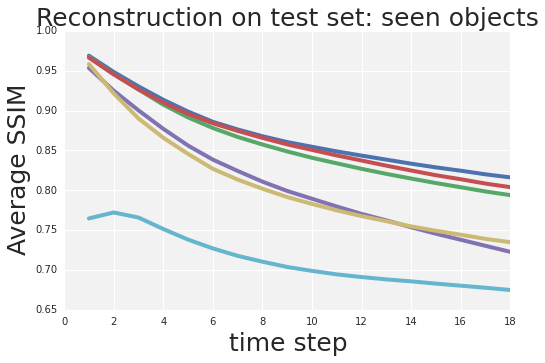}}
    %\put(1.33,0.4){\includegraphics[width=0.33\columnwidth]{motion_legend.png}}
    \put(1.0,0.0){\includegraphics[width=0.25\columnwidth]{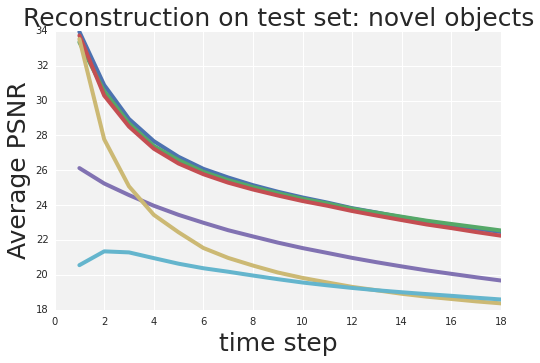}}
    \put(1.5,0.0){\includegraphics[width=0.25\columnwidth]{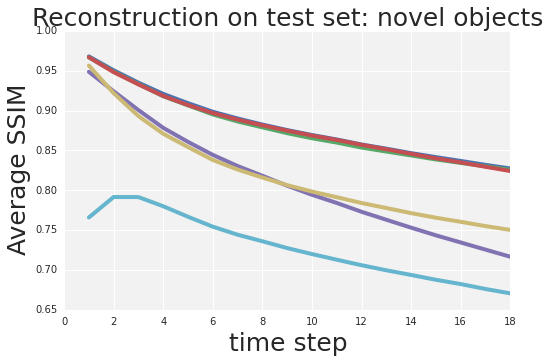}}
    \put(0.25,-0.05){\includegraphics[width=0.75\columnwidth]{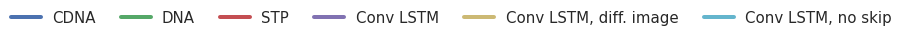}}

\end{picture}
\caption{Quantitative comparison to models which reconstruct rather than predict motion. Notice that on the novel objects test set, there is a larger gap between models which predict motion and those which reconstruct appearance.
\label{fig:moveablation}
}
\end{figure}

\begin{figure}
\setlength{\unitlength}{0.5\columnwidth}
\begin{picture}(1.99,0.3) \linethickness{0.5pt}
    \put(0.0,0.0){\includegraphics[width=0.25\columnwidth]{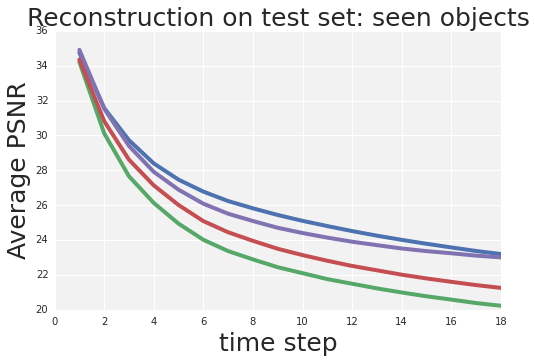}}
    \put(0.5,0.0){\includegraphics[width=0.25\columnwidth]{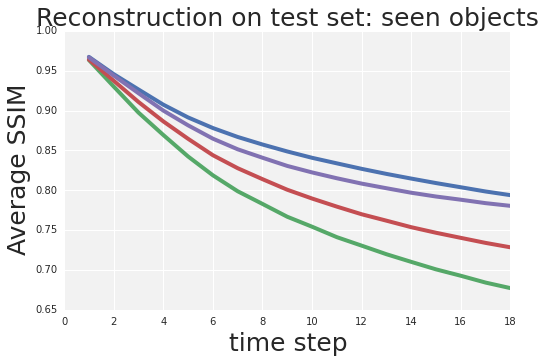}}
    \put(1.0,0.0){\includegraphics[width=0.25\columnwidth]{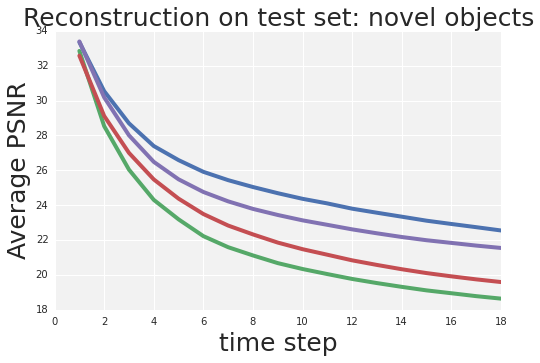}}
    \put(1.5,0.0){\includegraphics[width=0.25\columnwidth]{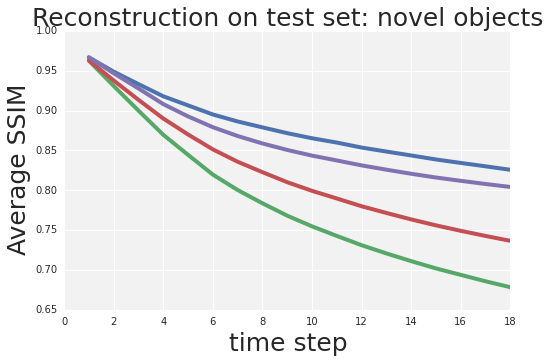}}
    \put(0.25,-0.05){\includegraphics[width=0.75\columnwidth]{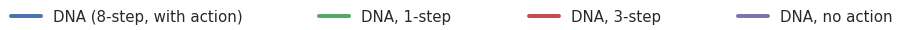}}

\end{picture}
\caption{Ablation of DNA involving not including the action, and different prediction horizons during training.
\label{fig:moveablation2}
}
\vspace{-0.12cm}
\end{figure}

\begin{figure}
\setlength{\unitlength}{0.5\columnwidth}
\begin{picture}(1.99,0.55) \linethickness{0.5pt}
    \put(-0.0,0.4){\includegraphics[width=0.5\columnwidth]{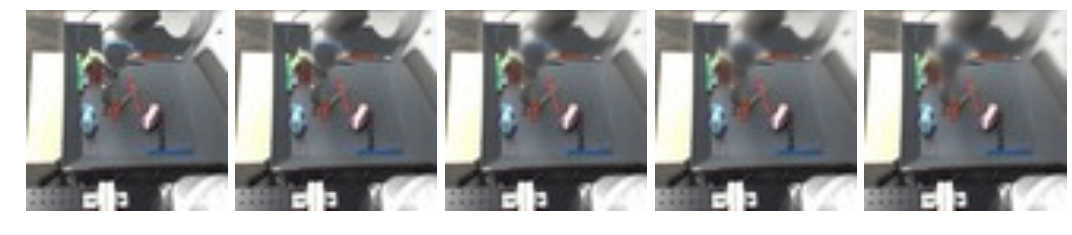}}
    \put(-0.0,0.2){\includegraphics[width=0.5\columnwidth]{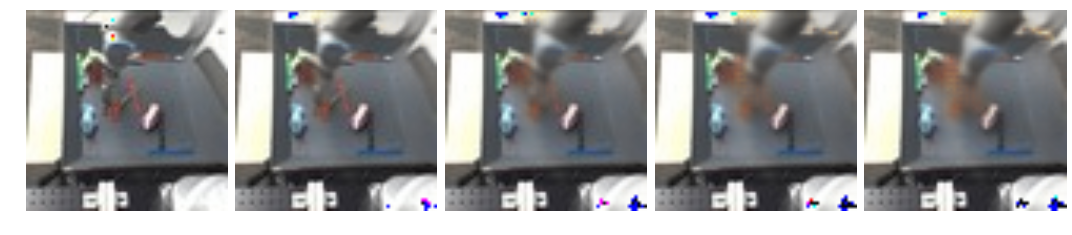}}
    \put(-0.0,0.0){\includegraphics[width=0.5\columnwidth]{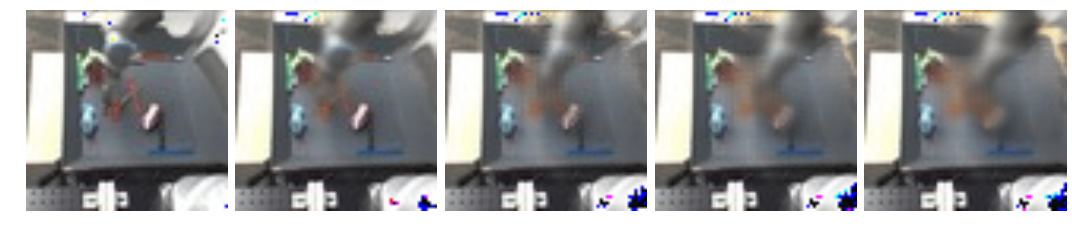}}

    \put(1.0,0.4){\includegraphics[width=0.5\columnwidth]{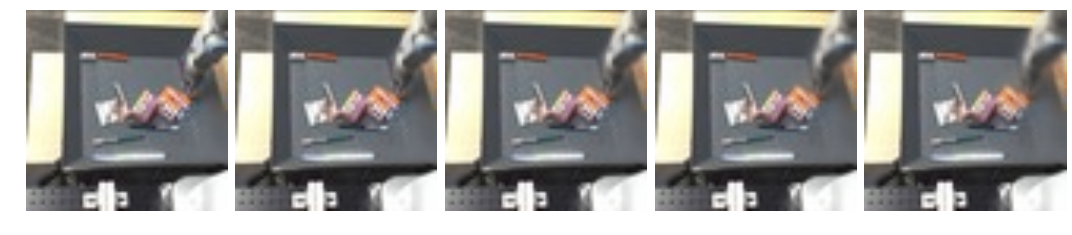}}
    \put(1.0,0.2){\includegraphics[width=0.5\columnwidth]{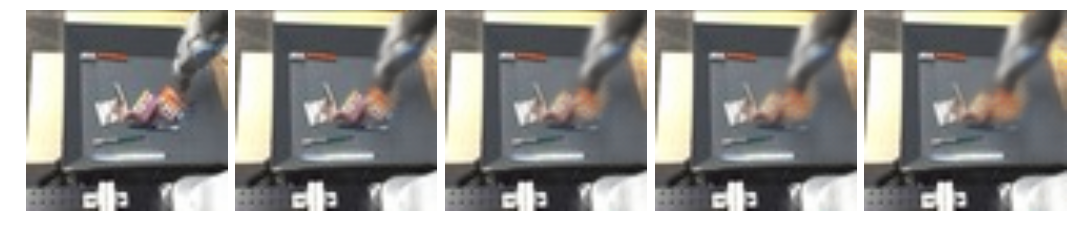}}
    \put(1.0,0.0){\includegraphics[width=0.5\columnwidth]{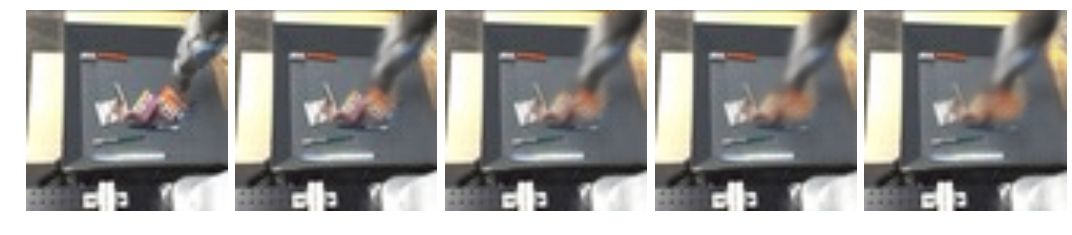}}

    \put(-0.03, 0.445){\rotatebox{90}{0 action}}
    \put(-0.03, 0.23){\rotatebox{90}{1x action}}
    \put(-0.03, 0.0){\rotatebox{90}{1.5x action}}

    \put(0.01, -0.02){t =}
    \put(0.1, -0.02){1}
    \put(0.3, -0.02){3}
    \put(0.5, -0.02){5}
    \put(0.7, -0.02){7}
    \put(0.9, -0.02){9}

    \put(1.1, -0.02){1}
    \put(1.3, -0.02){3}
    \put(1.5, -0.02){5}
    \put(1.7, -0.02){7}
    \put(1.9, -0.02){9}

\end{picture}
\caption{CDNA predictions from the same starting image, but different future actions, with objects
    \emph{not seen in the training set}. By row, the images show predicted future with
    zero action (stationary), the original action, and an action 150\% larger than the original.
    Note how the prediction shows no motion with zero action, and with a larger action, predicts more motion, including object motion.
\label{fig:arm}
}
\end{figure}
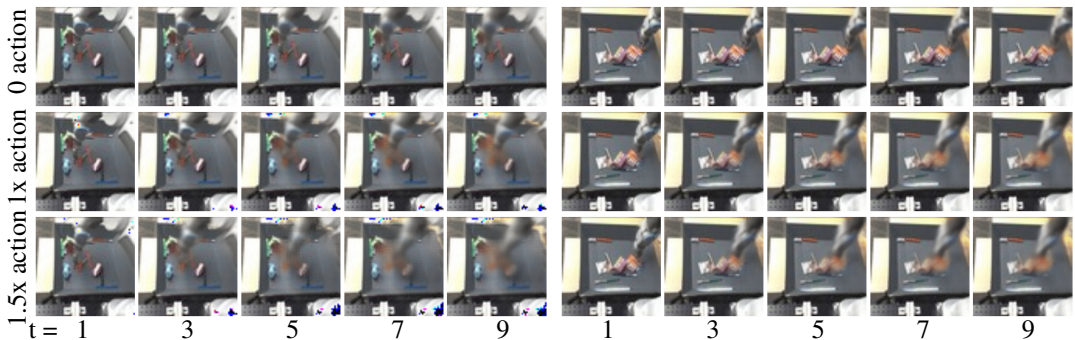

In Figure~\ref{fig:moveablation}, we compare to models with the same stacked convolutional LSTM architecture,
but that predict raw pixel values or the difference between previous and current frames. By explicitly
modeling pixel motion, our method outperforms these ablations. Note that the model without skip connections is most representative of the model by Xingjian et al.~\cite{xcwyw-clstm-15}.
We show a second ablation in
Figure~\ref{fig:moveablation2}, illustrating the benefit of training for longer horizons and from conditioning on the action of the robot.
Lastly, we show qualitative results in Figure~\ref{fig:arm} of changing the action of the arm to examine the model's predictions about possible futures.

For all of the models, the prediction quality degrades over time, as uncertainty increases further into the future. We use a mean-squared error objective, which
optimizes for the mean pixel
values. The model thus encodes uncertainty as blur.
Modeling this uncertainty directly through, for example, stochastic neural networks is an interesting direction for future
work. Note that prior video prediction methods have largely focused on
single-frame prediction, and most have not demonstrated prediction of multiple real-world RGB video frames in sequence.
Action-conditioned multi-frame prediction is a crucial ingredient in model-based planning, where the robot could mentally test the outcomes of various actions before picking
the best one for a given task.

\begin{comment}
\begin{figure}
\setlength{\unitlength}{0.5\columnwidth}
\begin{picture}(1.99,0.5) \linethickness{0.5pt}
\end{picture}
\caption{turk experiment
\label{fig:turk}
}
\end{figure}
\end{comment}

\subsection{Human motion prediction}

In addition to the action-conditioned prediction, we also evaluate our model on predicting future video without actions. We chose the Human3.6M dataset, which consists of
human actors performing various actions in a room.
% maybe mention why we chose this dataset (predict motion of entities)
We trained all models on $5$ of the human subjects, held out one subject for validation, and held out a different subject for the evaluations presented here. Thus, the
models have never seen this particular human subject or any subject wearing the same clothes.
We subsampled the video down to $10$ fps such that there was noticeable motion in the videos within reasonable time frames. Since the model is no longer conditioned
on actions, we fed in $10$ video frames and
trained the network to produce the next $10$ frames, corresponding to $1$ second each. Our evaluation measures performance up to $20$ timesteps into the future.

\begin{figure}
\setlength{\unitlength}{0.5\columnwidth}
\begin{picture}(1.99,1.05) \linethickness{0.5pt}
    \put(-0.0,0.9){\includegraphics[width=0.5\columnwidth]{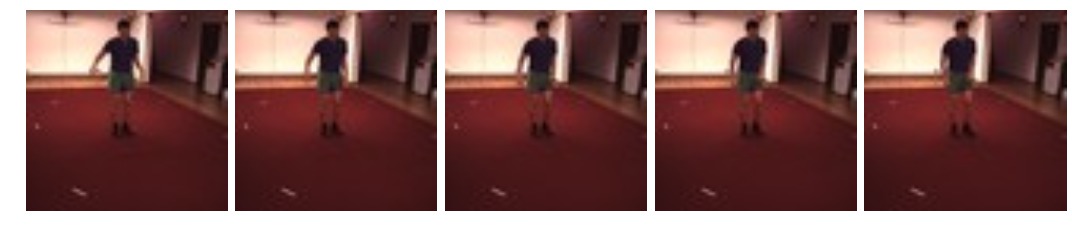}}
    \put(-0.0,0.7){\includegraphics[width=0.5\columnwidth]{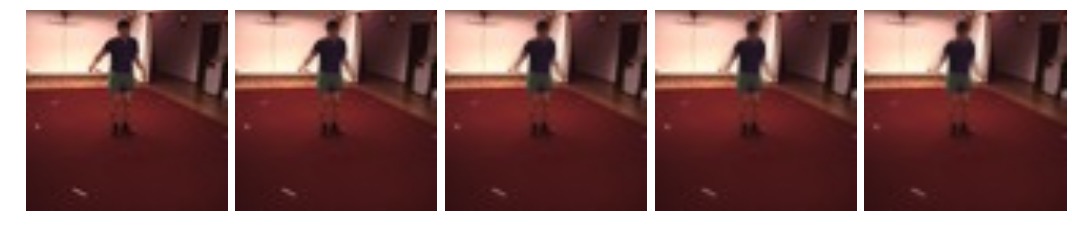}}
    \put(1.05,0.9){\includegraphics[width=0.5\columnwidth]{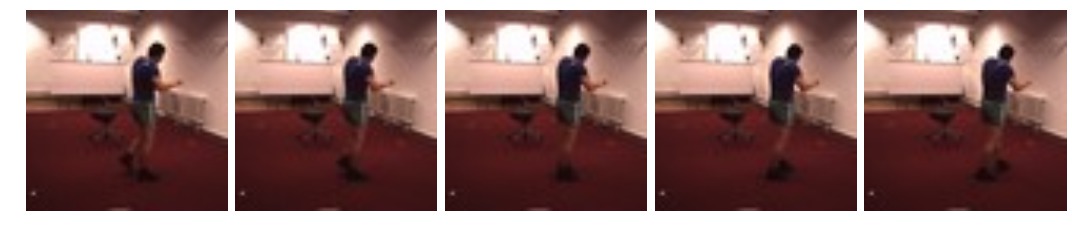}}
    \put(1.05,0.7){\includegraphics[width=0.5\columnwidth]{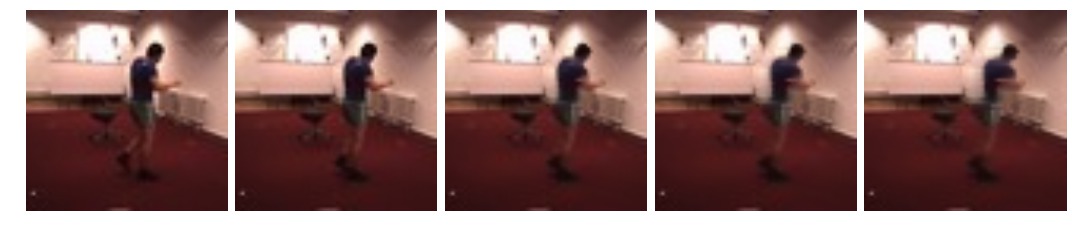}}
    \put(-0.0,0.5){\includegraphics[width=0.5\columnwidth]{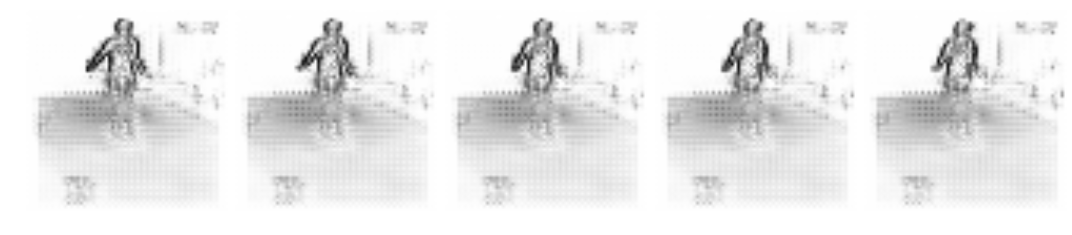}}
    \put(1.05,0.5){\includegraphics[width=0.5\columnwidth]{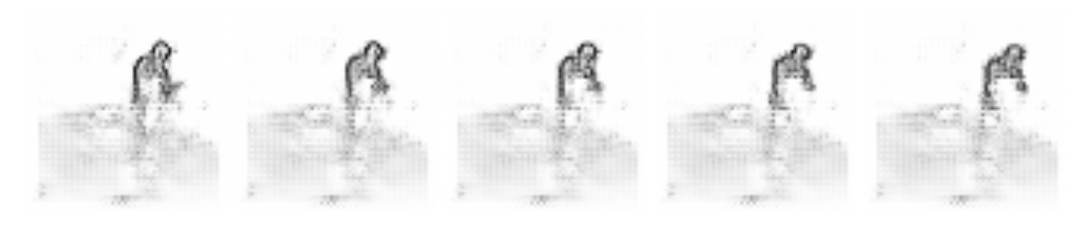}}

    \put(-0.04, 0.99){\rotatebox{90}{GT}}
    \put(-0.03, 0.79){\rotatebox{90}{STP}}
    \put(-0.03, 0.51){\rotatebox{90}{STP mask}}

    \put(0.0, 0.48){t =}
    \put(0.1, 0.48){1}
    \put(0.3, 0.48){4}
    \put(0.5, 0.48){7}
    \put(0.68, 0.48){10}
    \put(0.88, 0.48){13}

    \put(1.15, 0.48){1}
    \put(1.35, 0.48){4}
    \put(1.55, 0.48){7}
    \put(1.73, 0.48){10}
    \put(1.92, 0.48){13}

    \put(0.33,0.02){\includegraphics[width=0.33\columnwidth]{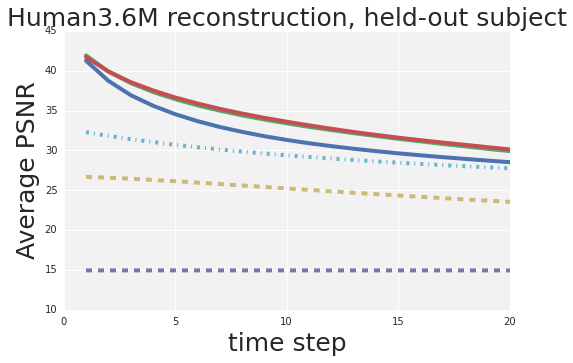}}
    \put(1.0,0.02){\includegraphics[width=0.33\columnwidth]{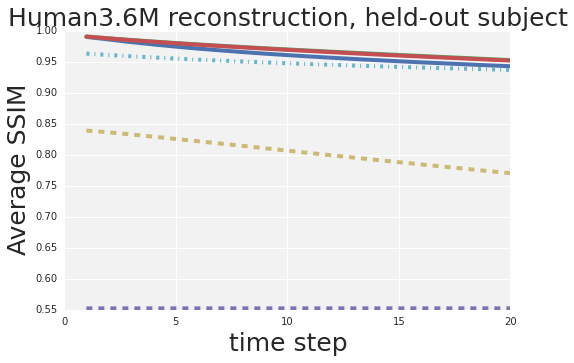}}
    %\put(1.33,0.02){\includegraphics[width=0.33\columnwidth]{flow_human.png}}
    \put(0.25,-0.03){\includegraphics[width=0.75\columnwidth]{comp_legend.png}}
\end{picture}
\caption{Quantitative and qualitative results on human motion video predictions with a \emph{held-out human subject}.
All recurrent models were trained for 10 future timesteps. %Note that the prediction quality degrades after 10 steps.
\label{fig:human}
}
\vspace{-0.1cm}
\end{figure}

The results in Figure~\ref{fig:human} show that our motion-predictive models quantitatively outperform prior methods, and qualitatively produce plausible
motions for at least $10$ timesteps, and start to degrade thereafter. We also show the masks predicted internally by the model for masking out the previous frame,
which we refer to as the background mask. These masks
illustrate that the model learns to segment the human subject in the image without any explicit supervision.

\section{Conclusion \& Future Directions}

In this work, we develop an action-conditioned video prediction model for interaction that incorporates appearance information in
previous frames with motion predicted by the model. To study unsupervised learning for interaction, we also present
a new video dataset with $59@000$ real robot interactions and $1.5$ million video frames. Our experiments show that, by learning to
transform pixels in the initial frame, our model can produce plausible video sequences more than $10$ time steps into the future, which corresponds to about one second.
In comparisons to prior methods, our method achieves the best results on a number of previous proposed metrics.

Predicting future object motion in the context of a physical interaction is a
key building block of an intelligent interactive system.
The kind of action-conditioned prediction of future video frames that we
demonstrate can allow an interactive agent, such as a robot, to imagine
different futures based on the available actions. Such a mechanism can be used to plan for actions to accomplish a particular goal, anticipate possible future problems
(e.g. in the context of an autonomous vehicle), and recognize interesting new phenomena in the context of exploration.
While our model directly predicts the motion of image pixels and naturally groups together pixels that belong to the same object and move together, it does not explicitly
extract an internal object-centric representation (e.g. as in \cite{ehwtkh-air-16}). Learning such a representation would be a promising future direction,
particularly for applying efficient reinforcement learning algorithms that might benefit from concise state representations.

%Another useful addition would be introducing stochasticity or uncertainty into the model, since
%physical interactions are inherently uncertain especially in the presence of occlusions.

\vspace{-0.1cm}
\subsubsection*{Acknowledgments}
We would like to thank Vincent Vanhoucke, Mrinal Kalakrishnan, Jon Barron, Deirdre Quillen, and our anonymous reviewers for helpful feedback and discussions.
We would also like to thank Peter Pastor for technical support with the robots.
%Use unnumbered third level headings for the acknowledgments. All
%acknowledgments go at the end of the paper. Do not include
%acknowledgments in the anonymized submission, only in the final paper.

\iffalse
\let\OLDthebibliography\thebibliography
\renewcommand\thebibliography[1]{
	\OLDthebibliography{#1}
	%\setlength{\parskip}{0pt}
	\setlength{\itemsep}{5pt plus 0.3ex}
}
\fi

\vspace{-0.1cm}
{\footnotesize
\bibliographystyle{abbrv}
\bibliography{references}
}

%References follow the acknowledgments. Use unnumbered first-level
%heading for the references. Any choice of citation style is acceptable
%as long as you are consistent. It is permissible to reduce the font
%size to \verb+small+ (9 point) when listing the references. {\bf
  %Remember that you can use a ninth page as long as it contains
  %\emph{only} cited references.}
%\medskip
%\small

\newpage

\appendix
\part*{Appendix}

\section{Data Collection Details}
\label{app:robot}

In this section, we discuss additional details of the data collection procedure.

The data was collected using $10$ 7-degree-of-freedom robot arms, and included the robot's joint angles, gripper pose, commanded gripper pose, measured torques,
and $640 \times 512$ RGB images captured from the robot's camera.
The images were center-cropped and downsampled to $64 \times 64$ for our experiments. We used smaller images for faster training.
In principle, the proposed methods should be able to handle larger images if desired. %Example images are shown in Figure~\ref{fig:data}.
The images and robot sensor readings were recorded at $10$ Hz, and the robot was commanded via impedance control in task space. In our experiments, the robot's
internal state was the gripper pose, and the action was the commanded gripper pose. Two test sets, each with $1@500$ recorded motions, were also collected.
Both used the same robot control method described above. The first test set used two different subsets of the objects pushed during training.
The second test set involved two subsets of objects, none of which were used during training.

During data collection, between ten and twenty random objects were placed in bins in front of each robot, and swapped out for new objects after approximately
$4@000$ randomized interactions. The robots were
programmed to repeatedly perform one of two different types of pushing motions: a random push or a randomized sweep to the middle. The sweep to the middle starts from a random
position on the outside border of the bin, and meandered randomly towards the middle. The sweep motion was designed to prevent objects from piling up on the edges of the bin.
Each type of motion lasted for approximately 3-5 seconds.
Between each motion, the arm was programmed to move out of the camera scene, and an image was recorded.

\section{Model Details}
\label{app:model}

\begin{figure}[b]
	\setlength{\unitlength}{0.5\columnwidth}
	\begin{picture}(1.99,0.7) \linethickness{0.5pt}
	\put(0.0,0.0){\includegraphics[width=1.0\columnwidth]{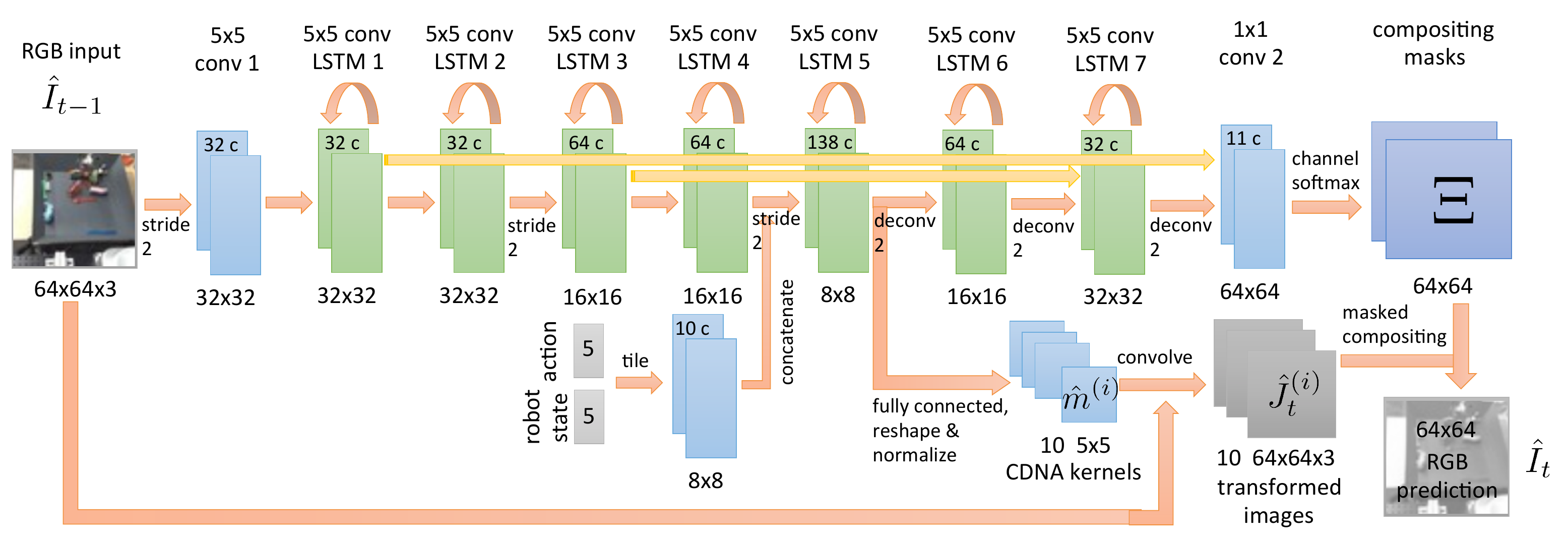}}
	\end{picture}
	\caption{Architecture of the DNA model. In contrast to the other models, the DNA model outputs the spatially-varying transformation kernels from the last layer,
	rather than the middle convolutional LSTM layer. The kernels are applied to transform the previous image, and the transformed image is composited with a 2-channel
	foreground-background mask.
		\label{fig:diagram_dna}
	}
\end{figure}

\begin{figure}
	\setlength{\unitlength}{0.5\columnwidth}
	\begin{picture}(1.99,0.7) \linethickness{0.5pt}
	\put(0.0,0.0){\includegraphics[width=1.0\columnwidth]{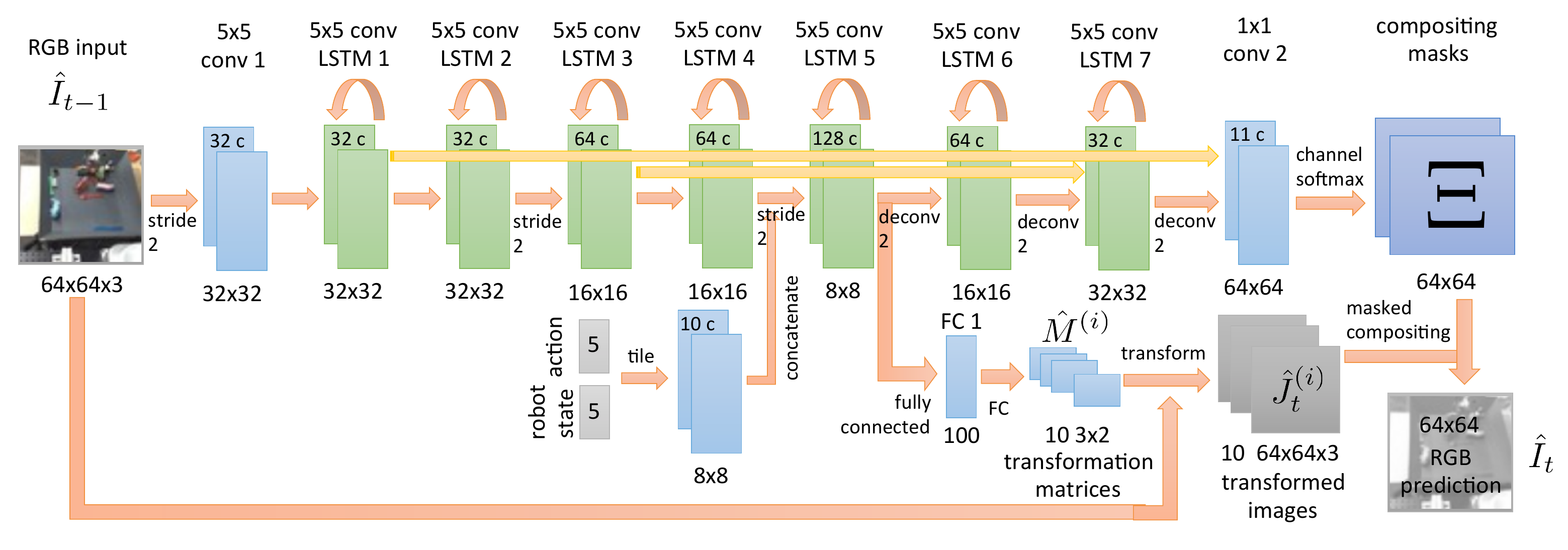}}
	\end{picture}
	\caption{Architecture of the STP model. This model is identical to the CDNA, with the only difference being that instead of outputting 10 transformation kernels,
	the model outputs 10 affine transformation matrices (with an additional fully connected layer). As with the CDNA,
	the transformations are each applied to the previous image, and the 10 resulting transformed images are composited by using a mask.
		\label{fig:diagram_stp}
	}
\end{figure}

In this section, we present additional details for each model. A diagram of the CDNA model is shown in the main paper in Figure~\ref{fig:diagram},
and additional diagrams for the DNA and STP models are shown in Figures~\ref{fig:diagram_dna} and \ref{fig:diagram_stp}, respectively.
Each model consists of a core trunk made up of one stride-2 $5\times 5$ convolution, followed by convolutional LSTMs.
Each of these LSTM layers has the weights arranged into $5\times 5$ convolutions, and the output of the preceding LSTM is fed directly into the next one.
LSTM layers 3 and 5 are preceded by stride 2 downsampling to reduce resolution, and LSTM layers 5, 6, and 7 are preceded by $2\times$ upsampling.
The end of the LSTM stack is followed by a $2\times$ upsampling stage and a final convolutional layer, which then outputs a full-resolution mask for compositing
the various transformed predictions (in the case of the CDNA and STP) and compositing against the static background (in the case of all models, including the DNA).
To preserve high-resolution information, skip connections are included from LSTM 1 to conv 2 and from LSTM 3 to LSTM 7.
The skip connections simply concatenate the skip layer activations and those of the preceding layer before sending them to the following layer (e.g. the input to LSTM
7 consists of the concatenation of LSTM 6 and LSTM 3).

In the case of the action-conditioned robot manipulation task, all three models also include as input the current state and action of the robot
(corresponding to gripper pose and gripper motion command).
This 10-dimensional vector is first tiled into a $8 \times 8$ response map with $10$ channels, and then concatenated, channel-wise, to the input of LSTM 5.
The next state is predicted linearly from the current state and action, though more sophisticated prediction models could be used for more complex systems.

The three models differ in the form of the transformation that is applied to the previous image.
The object-centric CDNA and STP models output the transformation parameters after LSTM 5.
In both cases, the output of LSTM 5 is flattened and linearly transformed, either directly into filter parameters in the case of the CDNA, or through one 100-unit hidden layer in the case of the STP.
There are $10$ CDNA filters, which are $5 \times 5$ in size and normalized to sum to 1 via a spatial softmax,
so that each filter represents a distribution over positions in the previous image from which a new pixel value can be obtained.
The STP parameters correspond to $10$ $3 \times 2$ affine transformation matrices. The transformations are applied to the preceding image to create
$10$ separate transformed images. The CDNA transformation corresponds to a convolution (though with the kernel being an output of the network),
while the STP transformation is an affine transformation.

The DNA model differs from the other two in that the transformation parameters are outputted at the last layer, in the same place as the mask.
This is because the DNA model outputs a transformation map as large as the entire image. For each image pixel, the model outputs a
$5 \times 5$ convolutional kernel that can be applied to the previous image to obtain a new pixel value, similarly to the CDNA model.
However, because the kernel is spatially-varying, this model is not equivalent to the CDNA. This transformation only produces one transformed image.

After transformation, the transformed image(s) and the previous image are composited together based on the mask.
The previous image is included as a static ``background'' image and, as shown in Section~\ref{experiments}, the mask on the background image indeed tends to pick out static parts of the scene.
The final image is formed by multiplying each transformed image and the background image by their mask values, and adding all of the masked images together.

\section{Additional Experimental Results}
\label{app:multiscale}

Here we show additional experimental results.

\begin{figure}
\setlength{\unitlength}{0.5\columnwidth}
\begin{picture}(1.99,0.35) \linethickness{0.5pt}
    \put(0.0,0.05){\includegraphics[width=0.25\columnwidth]{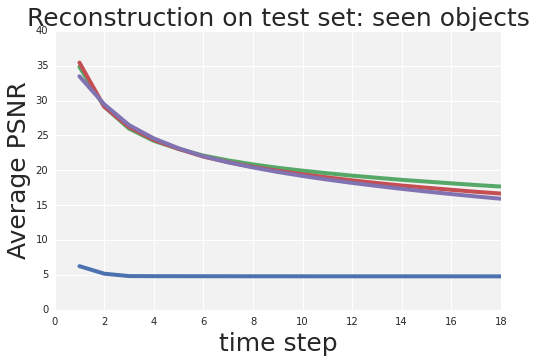}}
    \put(0.5,0.05){\includegraphics[width=0.25\columnwidth]{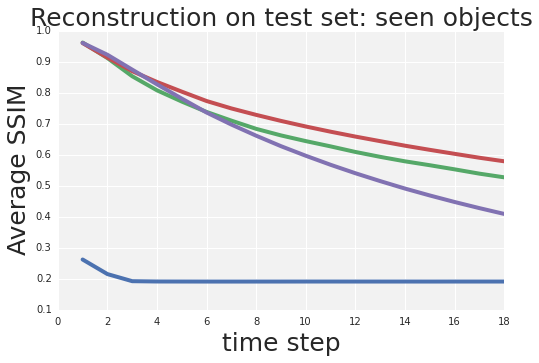}}
    \put(1.0,0.05){\includegraphics[width=0.25\columnwidth]{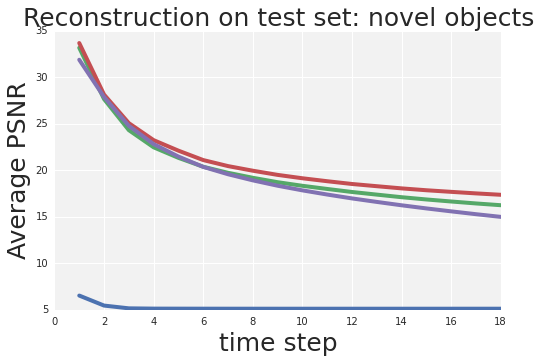}}
    \put(1.5,0.05){\includegraphics[width=0.25\columnwidth]{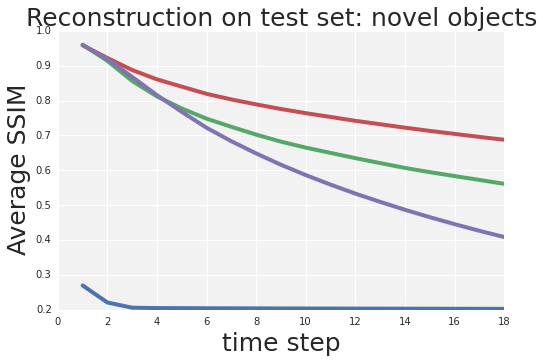}}
    \put(0.25,-0.0){\includegraphics[width=0.75\columnwidth]{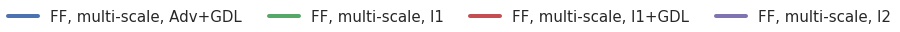}}

\end{picture}
\caption{Comparison of various losses using the feedforward multi-scale architecture from \cite{\nyu}. We were unable to get the adversarial objective to train desirably.
\label{fig:nyu}
}
\end{figure}

To evaluate the method of \cite{\nyu}, we trained the feedforward multiscale architecture using four of the proposed objectives. For the robot motion dataset,
we added the actions to the
architecture at each scale via tiling. We were unable to successfully train the model with an adversarial loss, and, as shown
in Figure~\ref{fig:nyu}, the model trained GDL+$l_1$ performed the best on the robot dataset. Thus, we presented the results of this model in the main evaluation.

\end{document}